\documentclass{article}

\usepackage{microtype}
\usepackage{graphicx}
\usepackage{subfigure}
\usepackage{booktabs} 
\usepackage{enumerate}  
\usepackage{amsmath}
\usepackage{subfigmat}
\usepackage{changepage}
\usepackage{float}
\usepackage{hyperref}

\usepackage[accepted]{icml2018}

\begin{document}

\twocolumn[
\icmltitle{Boundary Optimizing Network (BON)}




\icmlsetsymbol{equal}{*}

\begin{icmlauthorlist}
\icmlauthor{Marco Singh}{equal,corti}
\icmlauthor{Akshay Pai}{equal,diku,bmq}
\end{icmlauthorlist}

\icmlaffiliation{corti}{Corti, Denmark}
\icmlaffiliation{diku}{University of Copenhagen, Department of Computer Science, Denmark}
\icmlaffiliation{bmq}{Biomediq A/S, Denmark}

\icmlcorrespondingauthor{Marco Singh}{ms@cortilabs.com}
\icmlcorrespondingauthor{Akshay Pai}{akshay@biomediq.com}
\icmlkeywords{Machine Learning, Deep Learning, Artificial Intelligence, Boundary Optimization, Supervised Learning, Generative Network, Data Augmentation, Noise in Deep Learning, Noisy Networks}

\vskip 0.3in
]


\printAffiliationsAndNotice{\icmlEqualContribution} 

\begin{abstract}
Despite all the success that deep neural networks have seen in classifying certain datasets, the challenge of finding optimal solutions that generalize still remains. In this paper, we propose the Boundary Optimizing Network (BON), a new approach to generalization for deep neural networks when used for supervised learning. Given a classification network, we propose to use a collaborative generative network that produces new synthetic data points in the form of perturbations of original data points. In this way, we create a data support around each original data point which prevents decision boundaries from passing too close to the original data points, i.e. prevents overfitting. We show that BON improves convergence on CIFAR-10 using the state-of-the-art Densenet. We do however observe that the generative network suffers from catastrophic forgetting during training, and we therefore propose to use a variation of Memory Aware Synapses to optimize the generative network (called BON++). On the Iris dataset, we visualize the effect of BON++ when the generator does not suffer from catastrophic forgetting and conclude that the approach has the potential to create better boundaries in a higher dimensional space. 
\end{abstract}

\section{Introduction}
Despite the significant success that deep neural networks (DNNs) have seen in recent times, the generalization abilities remain questionable \cite{2016arXiv161103530Z}. The lack of generalization is due to overparameterization of the network and subsequent overfitting during optimization. Regularization techniques such as restricting the magnitude of the parameter values, e.g. $l^2$-norm, or injecting noise during training, e.g. dropout \cite{2012arXiv1207.0580H}, are employed to increase the ability of the network to generalize to unseen data. 
Explicit regularization techniques restrict the expressiveness of the network, and implicit regularization methods such a noise corruption \cite{2017arXiv171005179N}, \cite{ICLR2017} need very careful choices of noise forms. 

In this paper we propose Boundary Optimizing Networks (BONs) that regularize neural networks by introducing synthetic data points derived from the original data. In particular, we employ a generative network that works in a collaborative fashion and generates noisy data points which are then used to train a neural network. In this way, we create a data support around each original data point which prevents decision boundaries from passing too close to the original data points, i.e. prevents overfitting. As conjectured in \citet{2016arXiv161200138R}, generalization of a DNN is diminished due to diminishing gradients of correctly classified samples. In contrast, BON works in the opposite direction - the gradients diminish in misclassified samples relative to correctly classified samples which creates significant data support around correctly classified data points. To prevent catastrophic forgetting during training, we propose to use a variation of Memory Aware Synapses to optimize the generative networks - this variation is referred to as BON++ in the rest of the paper. On the Iris dataset, we show that the BON algorithm successfully creates data support in densely populated areas and hence prevents the classifier from overfitting to outliers.

Similarities may be drawn to two famous versions of generative networks: a) Generative Adversarial Networks (GANs) \cite{2014arXiv1406.2661G}, and b) Generative Adversarial Perturbations (GAPs) \cite{2013arXiv1312.6199S}. The differences from BONs are as follows:
\begin{enumerate}[(i)]
\item In contrast to GAN, BON creates learned augmentations of data points. 
\item The loss function of the generator is pushing the synthetic data point closer to the real data point rather than trying to fool $D$, hence supporting $D$ instead of fooling $D$.
\item Unlike GAPs, the perturbations are not necessarily imperceptible. In addition, the perturbations are designed to keep the noisy datapoint around the original datapoint in order to help optimization.
\end{enumerate}

\section{Related Work}
The proposed work is an overlap between the areas of data augmentation and regularization by noise. Both relate to each other in the sense that both areas address the issue of generalization of a DNN. Data augmentation involves implicitly coupling a synthetic point lying in the vicinity of the original data point it is derived from \cite{2017arXiv171204621P}. Each synthetic data point is typically generated by varying the attributes of the data point, for example by rotations. More advanced approached (generative models) aim at using DNNs to approximate the data distributions and subsequently finding new data points by sampling from the distribution \cite{2017arXiv170901643R} \cite{2016arXiv161101331S} \cite{NIPS2001_2020}. 

On the other hand, regularization by noise works by injecting noise of a particular form into the weights of a network during optimization. The most popular way of noise-based regularization is dropout. Here, certain hidden units in a neural network are randomly turned off by multiplying noise sampled from a Bernoulli distribution. Some other approaches directly modify the weights of neural networks by injecting controlled noise \cite{pmlr-v28-wan13} \cite{2015arXiv150602557K}. Noise in the form of an ensemble of networks is another form of noise injection during the training of a DNN \cite{46511} \cite{2016arXiv160309382H}. Our work changes the injection of noise, by perturbing data points with controlled noise. The controlled noise is perturbed in a specific way such as to obtain better decision boundaries for the classifier.

The paper is organized as follows: in the next section, we will introduce BON and the formalisms around it. Following this, we will apply BON on CIFAR-10 \cite{cifar10} using state-of-the-art Densenet \cite{2016arXiv160806993H}. We then continue with a constructed dataset to visually show what is happening and furthermore develop the BON++ to fight catastrophic forgetting. BON++ is then tested on the Iris dataset to show that the proposed algorithm indeed does prevent overfitting from happening 

\section{Boundary Optimizing Network (BON)}

The idea behind BONs is that we couple a synthetic data point to each of the real data points in a given dataset. The synthetic data point is created through a generative network, $G$, which takes as input the real data point, $x$, and outputs a synthetic data point $\hat x$ . $G$ can be any neural network which takes as input the real data point and creates distortions to this input, i.e. the output dimensionality equals that of the input. In fact the only restriction on $G$ for now is that it should be a network with enough parameters to be able to produce distortions on all data points in the data set. 

The main classification network, $D$, can be any model best suited for the given dataset at hand. The BON approach helps defining new data points in order to make $D$ create boundaries that will generalize better, hence it is invariant to the actual structure of $D$. 

\subsection{Algorithm}

First we denote a synthetic data point $G(x^{(i)}) = \hat x^{(i)}$ as $\hat x^{(i)}_M$ if  
$$\hat c = \operatorname*{arg\,max}_j D(y=j \vert \hat x^{(i)}) \neq y^{(i)}$$


This simply means that we place a subscript of $M$ if the classification network, $D$, misclassified the synthetic data point. 

\begin{figure*}[ht]
\vskip 0.2in
\begin{center}
\centerline{\includegraphics[width=14cm]{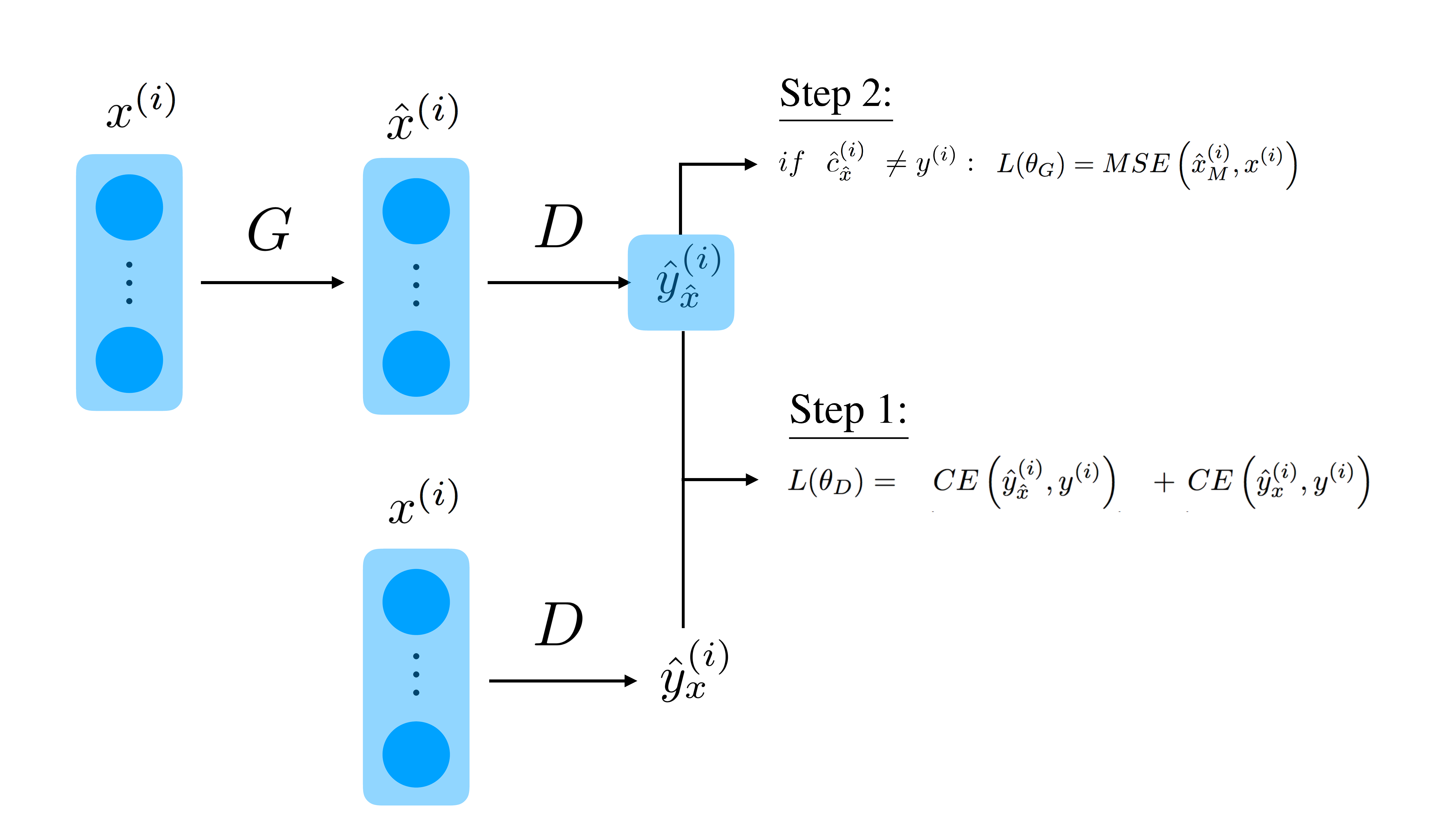}}
\caption{Illustration of BON algorithm: (i) Compute the synthetic data point $G(x)=\hat x$ (ii) Update $D$ using the cross-entropy loss of both the real and synthetic data point (iii) Update $G$ using the MSE loss of the real and synthetic data point - only if the synthetic data point is misclassified by $D$}
\label{graph_bon}
\end{center}
\vskip -0.2in
\end{figure*}

The algorithm starts by feeding a real data point through $G$ to create a synthetic data point $\hat x$ . Both the real and the synthetic data point are fed to $D$ with \textbf{the same} labels. The parameters of $D$ are then updated using the accumulated gradients from both the real and synthetic data point. 
Assuming the cross-entropy (CE) loss function, we minimize:

\begin{equation}
L(\theta_D)  = \underbrace{CE\left(\hat y^{(i)}_{\hat x}, y^{(i)}\right)}_{\text{loss from synthetic data point}} + \underbrace{CE\left(\hat y^{(i)}_{x}, y^{(i)}\right)}_{\text{loss from real data point}}
\end{equation}
where $\hat y$ denotes the prediction of $D$.

After updating $D$ we evaluate the performance of $D$ on the synthetic data point; if $\hat x$ is misclassified we have most likely altered the real data point too much, hence we minimize the mean squared error between the synthetic data point and the real data point it was created from when we update the parameters of $G$: 

\begin{equation}
L(\theta_G)  = MSE\left(\hat x^{(i)}_M, x^{(i)}\right)
\end{equation}If the evaluation shows that $D$ correctly classified $\hat x$ then we do \textit{not} update the parameters of $G$. If we correctly classify, we have added just enough noise to the original data point, and we hope to have created a supporting data point for a better generalizing decision boundary.

The computational structure for a single generator is illustrated in Figure~\ref{graph_bon} and the actual algorithm is shown in Algorithm~\ref{alg:bon}.

 \begin{algorithm}[h!]
   \caption{BON}
   \label{alg:bon}
\begin{algorithmic}
   \STATE {\bfseries Input:} dataset $(x^{(i)}, y^{(i)}), i \in \{1,2,...,m\}$
   \STATE
   \FOR{$n=1$ {\bfseries to} $\# \ epochs$}
	\FOR{$i=1$ {\bfseries to} $m$}
	\FOR{$k=1$ {\bfseries to} $K$}   
    \STATE
   \STATE $\bullet$ Create synthetic data point $G_k(x^{(i)}) = \hat x^{(i)}_k$   
   \STATE
   \STATE $\bullet$ Compute prediction of synthetic data point: \\
   \STATE
	\STATE   $\quad \quad \quad \quad \quad \quad D(\hat x^{(i)}_k) = \hat y^{(i)}_{\hat x_k}$      
   \STATE
      \ENDFOR
   \STATE
	\STATE $\bullet$ Compute prediction of real data point:
    \STATE
    $\quad \quad \quad \quad \quad \quad D(x^{(i)}) = \hat y^{(i)}_{x}$ 
    \STATE
	\STATE $\bullet$ Compute loss for $D$: \\
    \STATE    
   \STATE \resizebox{.9\hsize}{!}{$L(\theta_D)  = \sum_{k} CE\left[D\left(G_k\left(x^{(i)}\right)\right),  y^{(i)}\right]+ CE\left[\hat y^{(i)}_{x},  y^{(i)}\right]$}   
   \STATE    
	\STATE $\bullet$ Update $\theta_D$ to minimize $L(\theta_D)$ 
   \STATE
\IF{$\hat c \neq y^{(i)}$}
   \STATE
	\FOR{$k=1$ {\bfseries to} $K$}   
    \STATE        
	\STATE $\bullet$ Compute loss for $G_k$:\\
    \STATE    
   \STATE  \quad $L(\theta_{G_k}) =MSE\left[G_k\left(x^{(i)}\right), x^{(i)}\right]$ 
   \STATE
	\STATE $\bullet$ Update $\theta_{G_k}$ to minimize $L(\theta_{G_k})$    
   \STATE
   \ENDFOR
	\ENDIF
   \ENDFOR
   \ENDFOR
\end{algorithmic}
\end{algorithm}

So far we have only mentioned a single generative network $G$, but to create more data support, we will use a population of different generator networks $G_k, k \in \{1,2,...,K\}$. 
Each $G_k$ has the task mentioned so far, i.e. it will be creating a synthetic data point for each of the data points in the training data. In this way we create several synthetic data points for each real data point and hence create a cloud of data points supporting the decision boundary.

\section{Experiment: BON}
We now show how $G$ is improving training by using Densenet on CIFAR-10. We show the convergence using BON and compare to the original version without BON.

We then construct a fake data set to visualize how the synthetic data points are created and to explain the behavior of BON on CIFAR-10.

\subsection{CIFAR-10}
In this experiment we use Densenet with a depth of $100$ and a growth rate of $12$ on CIFAR-10+, i.e. CIFAR-10 with the same data augmentation scheme as presented in the original Densenet paper \cite{2016arXiv160806993H}. We use a single generator, i.e. $K=1$, with $5$ layers of 2D convolutions with an increasing number of channels except from the last layer which collapses the number of channels to $3$ as the original input image. We use a kernel size of $3$ and stride of $1$ and furthermore use Batch Normalization \cite{2015arXiv150203167I} and ReLU activations after each convolution layer except the last one.

\begin{figure}[t!]

\centering
\includegraphics[trim= 0cm 0cm 0cm 0cm,scale=0.3]{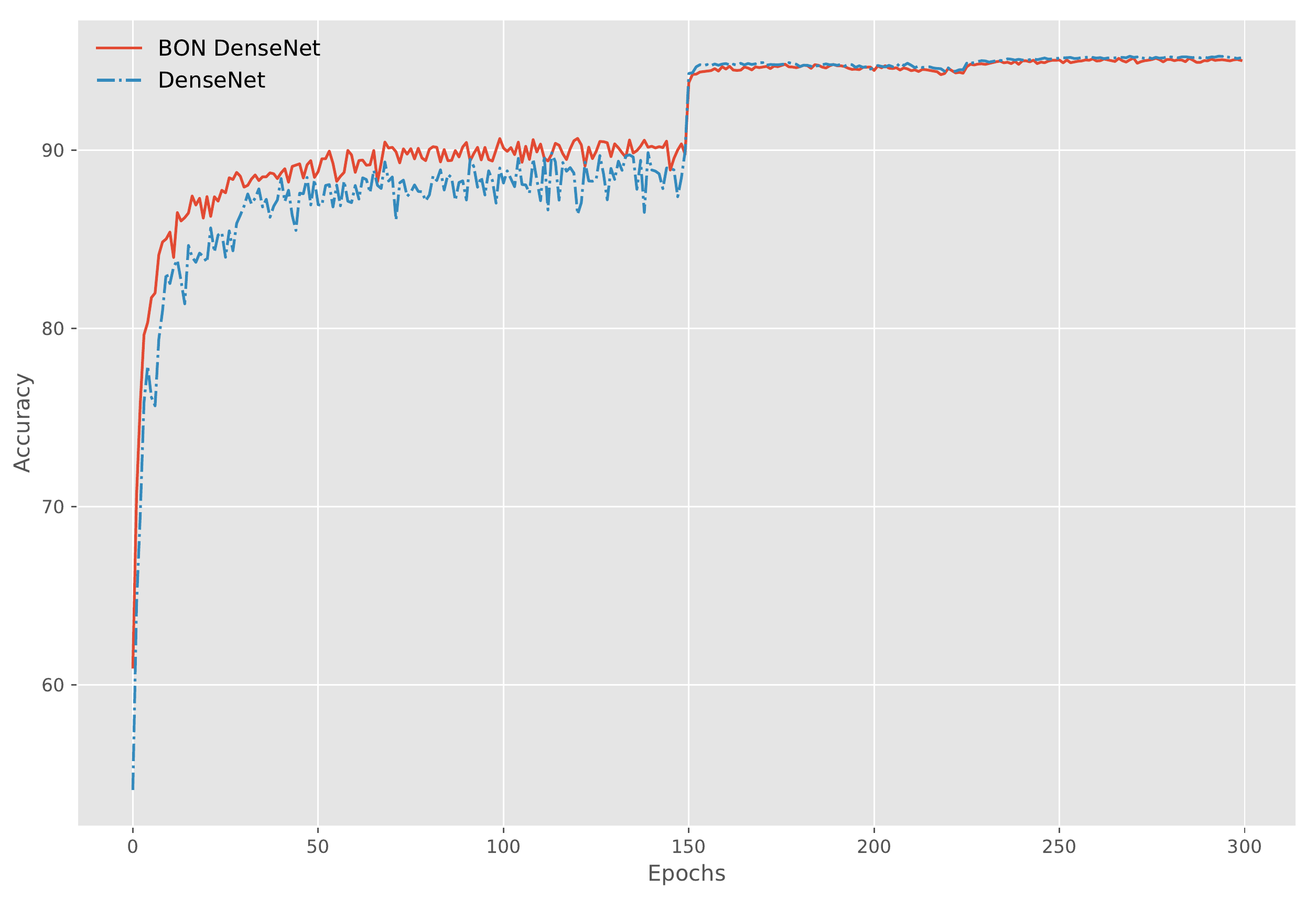}\hspace{0.1em}
\caption{BON algorithm using Densenet on CIFAR-10+}
\label{DensenetC10+}

\end{figure}

\begin{figure}[t!]

\centering
\includegraphics[trim= 0cm 0cm 0cm 0cm,scale=0.3]{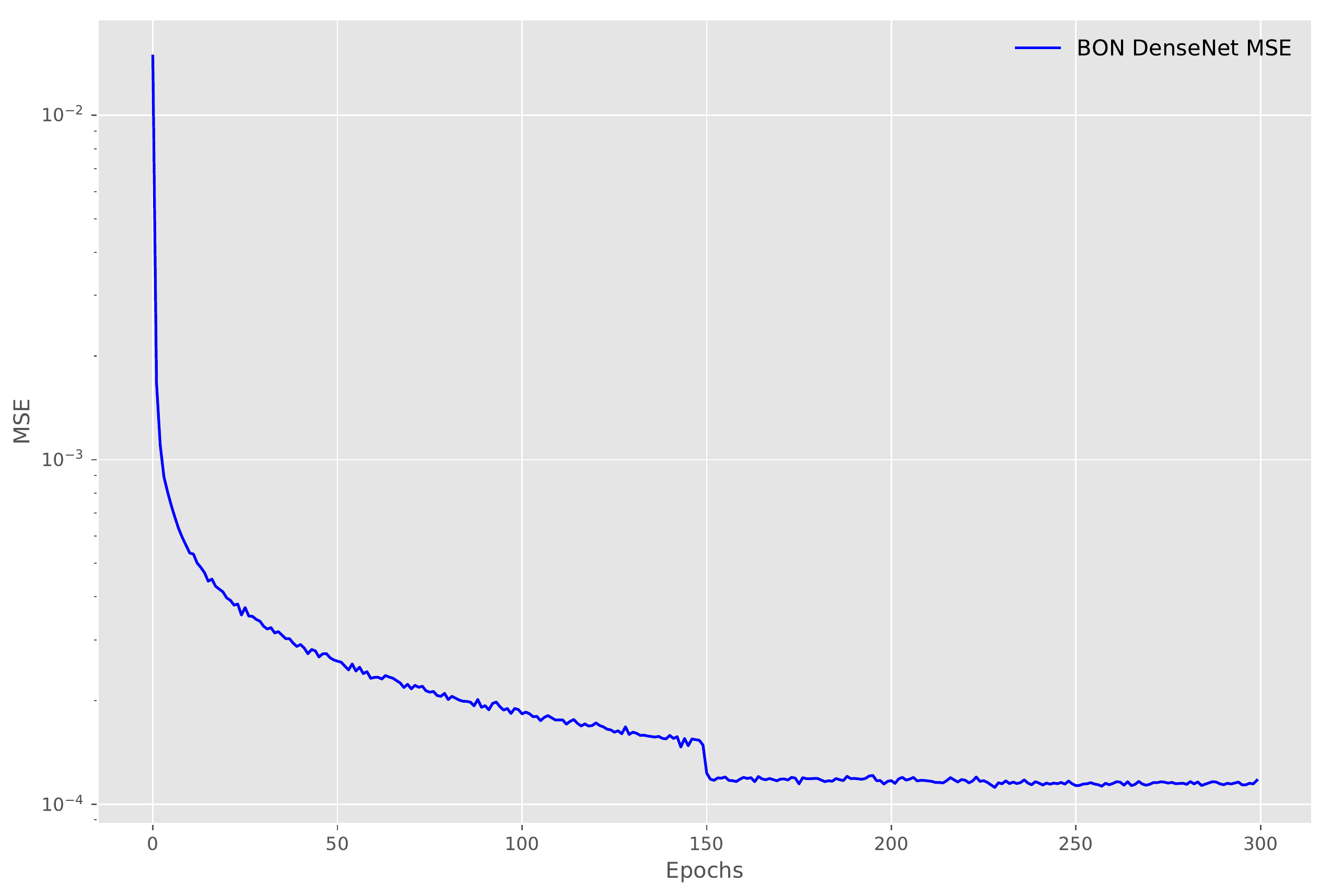}\hspace{0.1em}
\caption{Mean squared error of synthetic and real data points during training}
\label{mse}

\end{figure}

Figure~\ref{DensenetC10+} shows the accuracy as training progresses. It is clear that the BON approach helps convergence happen faster. We use the exact same learning rate schedule as in the Densenet paper, i.e. decrease the learning rate at epoch $150$ and $225$.
Although the faster convergence is a desirable feature, it is not the main objective of the BON. We are interested in making better boundaries and hence we would have expected a higher accuracy at the end of training. This does not happen, instead the BON approach produces equivalent accuracies to Densenet without BON. To get an understanding of why this happens, we have plotted the mean squared error (MSE) between \emph{all} the synthetic data points and their real equivalent as training progresses. Figure~\ref{mse} shows this. We would expect the MSE to decrease very slowly since we are no longer pushing the synthetic data points closer to their real equivalent as soon as they are correctly classified by $D$. Since Densenet classifies most samples correctly (it achieves an accuracy of $80\%$ after a few epochs) we would only expect the marginal change in the MSE for all samples to be small. This is however not the case, instead the MSE collapses to almost $0$ for all the samples, including the ones that were correctly classified early in training. 

This happens because the network only sees the misclassified samples for many epochs, hence the parameter update of $G$ is only using information from a few misclassified samples which are propagated closer and closer to their real equivalent. As this is happening, $G$ collapses to the indicator function and hence learns to replicate a real data point. This catastrophic forgetting is preventing BON from working as intended in its current state.

To visualize the behavior we construct a fake data set in the next section. It then becomes clear that the misclassified samples are causing catastrophic forgetting, and we then proceed with a cure by developing BON++.

\subsection{Constructed Experiment}
In this experiment we use a feedforward neural network with $3$ hidden layers and an output layer for $D$. The first $2$ hidden layers have $100$ neurons, the third layer has $50$ neurons and the output has $2$ neurons to match the number of classes. We use ReLU activations.

We create $10$ generators, i.e. $K=10$. For each $G_k$ we use $3$ hidden layers with $50$ neurons in each and a linear output layer with $2$ neurons to match the input dimensionality.
$G$ also make use of ReLU activations and the output layer has no activation as we want to create synthetic data points in the same range as the real data points.

The data consists of $2$ classes and have $2$ features. This is chosen such that we can easily visualize the decision boundary and the evolution of the synthetic data points during training. 

\begin{figure*}[t!]

\centering
\includegraphics[trim= 0.5cm 0cm 0cm 2cm,scale=0.27]{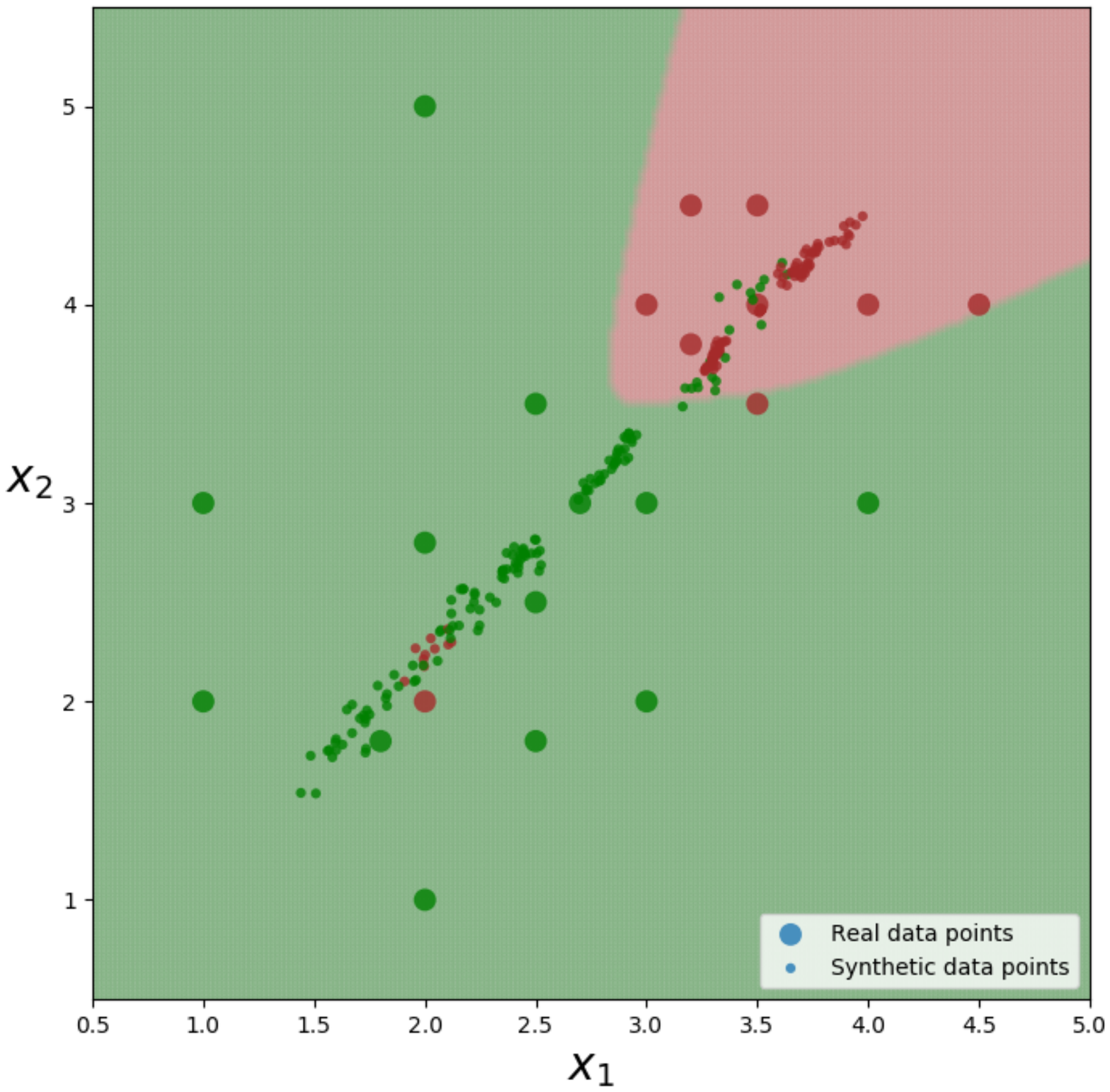}\hspace{0.1em}
\includegraphics[trim= 0.5cm 0cm 0cm 2cm,scale=0.27]{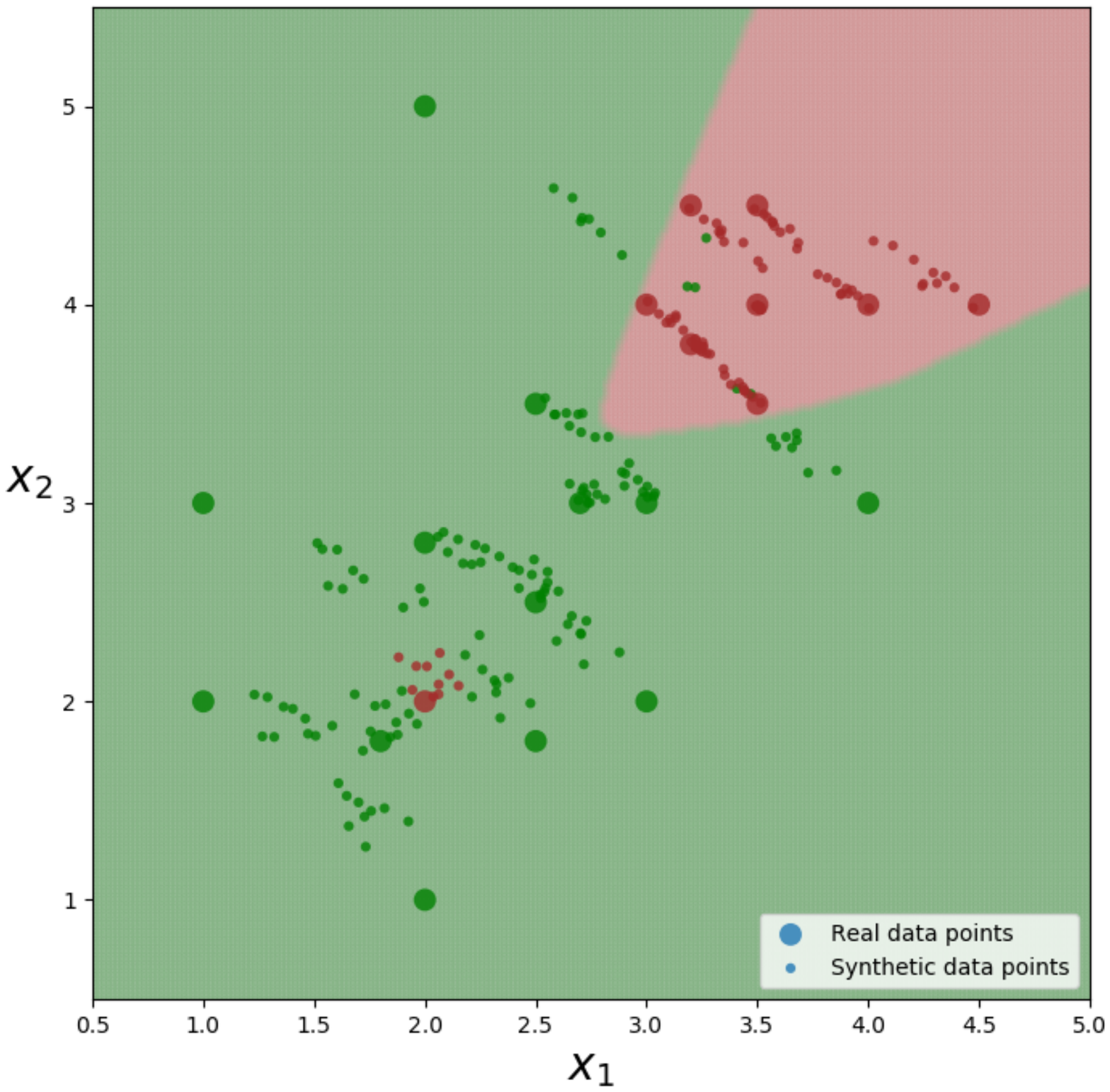}\hspace{0.1em}
\includegraphics[trim= 0.5cm 0cm 0cm 2cm,scale=0.27]{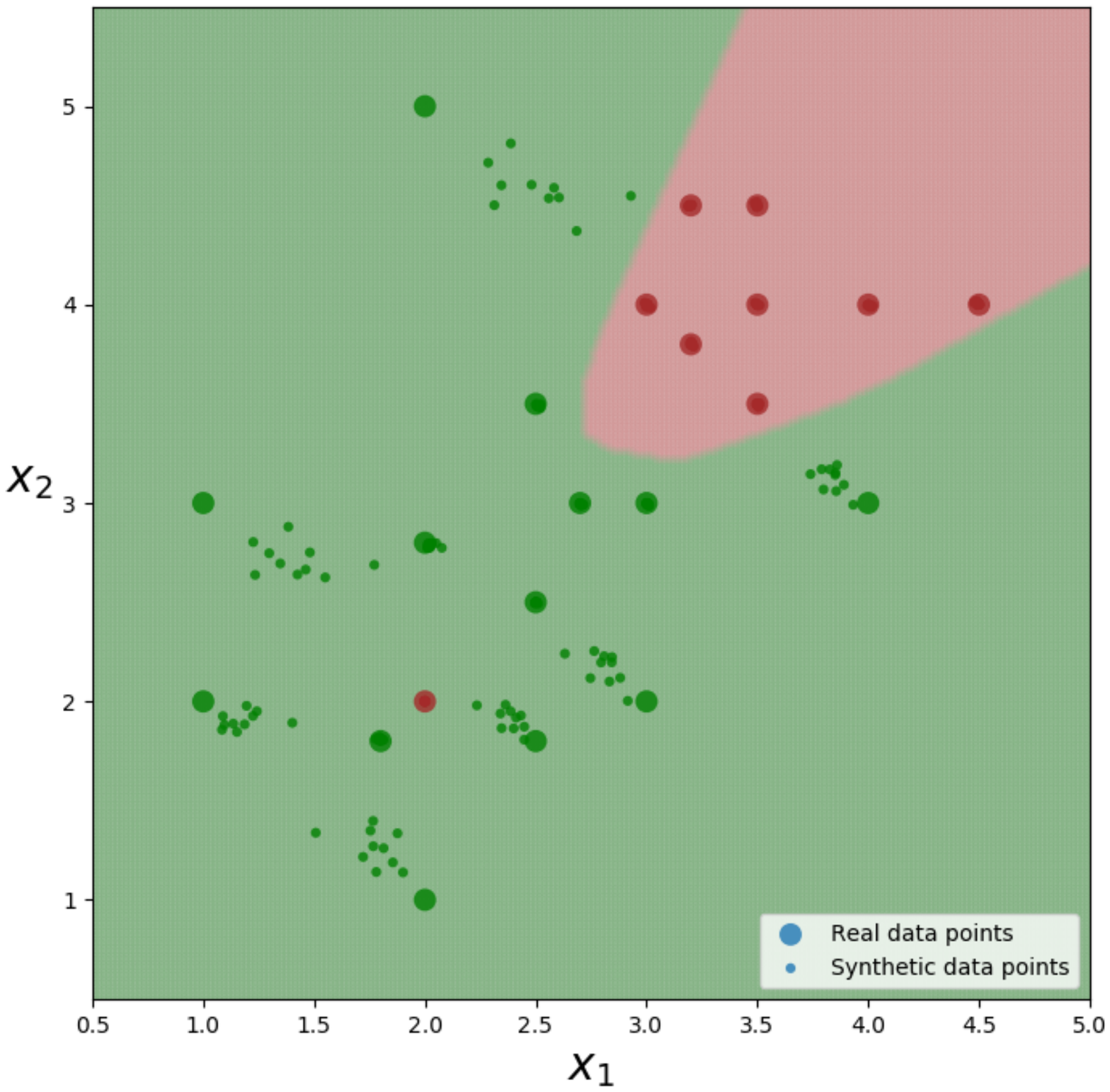}
\vspace*{-1.5cm}
\caption{BON algorithm on constructed data set. Real data points (large points) and synthetic data points (small points) are plotted together with the decision boundary for $D$. Left: epoch $50$, center: epoch $150$, right: epoch $500$}
\label{simple_bon}

\end{figure*}

\begin{figure*}[t!]

\centering
\includegraphics[trim= 0.5cm 0cm 0cm 1.2cm,scale=0.28]{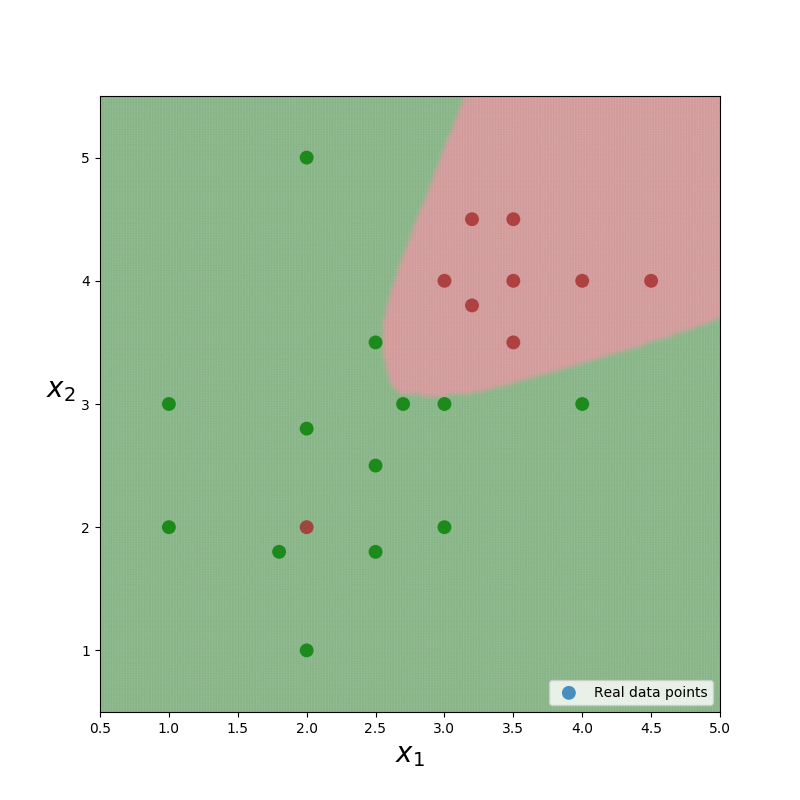}\hspace{0.1em}
\includegraphics[trim= 0.5cm 0cm 0cm 1.2cm,scale=0.28]{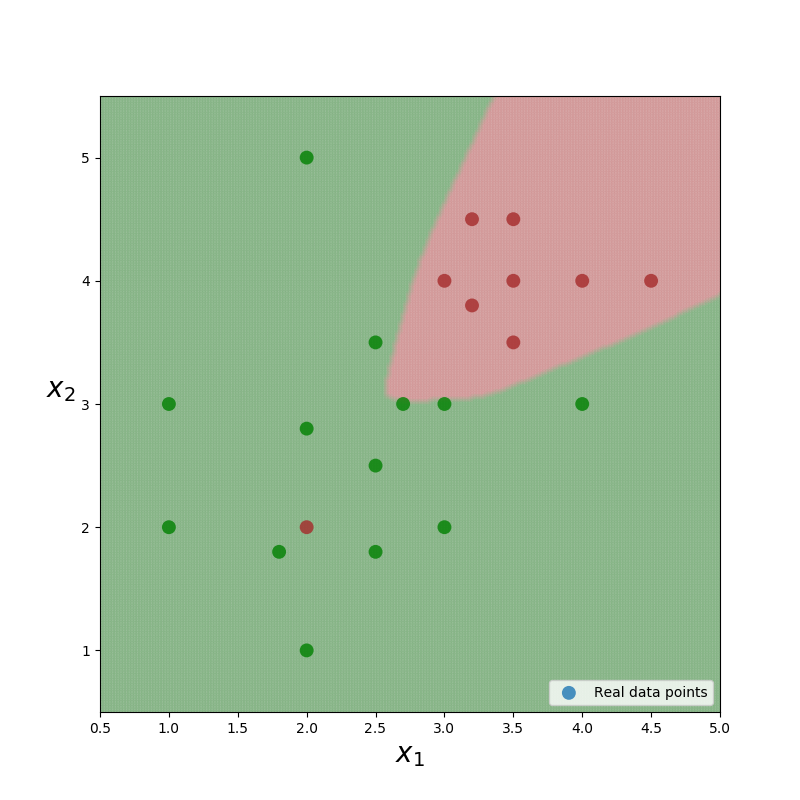}\hspace{0.1em}
\includegraphics[trim= 0.5cm 0cm 0cm 1.2cm,scale=0.28]{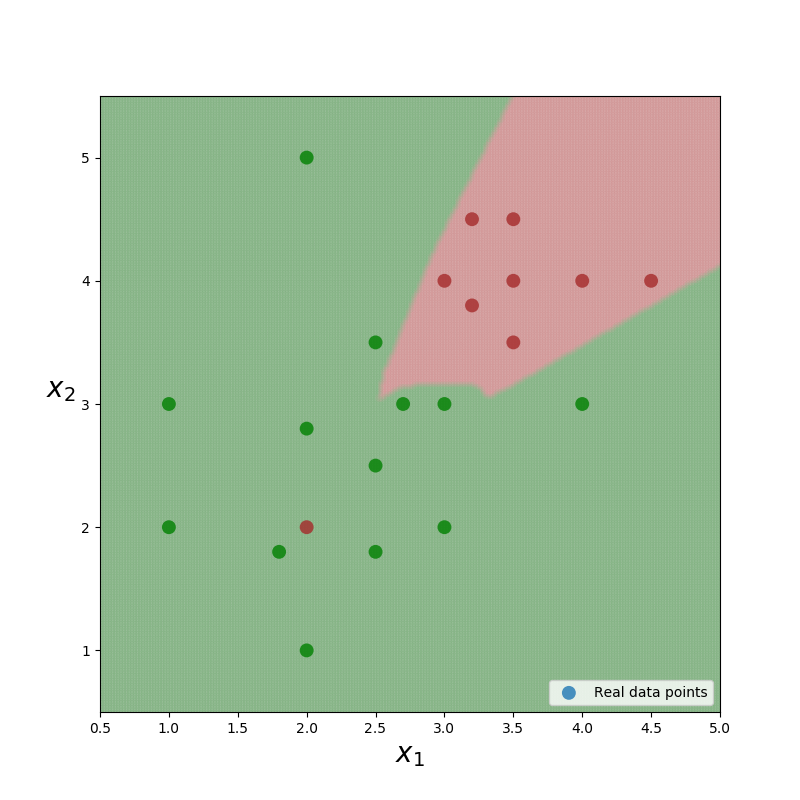}
\vspace*{-0.5cm}
\caption{Feedforward neural network, $D$, trained on constructed dataset. Left: epoch $50$, center: epoch $150$, right: epoch $500$}
\label{simple_bon_pure}

\end{figure*}

Figure~\ref{simple_bon} shows the simple example. After $150$ epochs, the synthetic data points have created exactly the support one would wish for. The synthetic data points are surrounding the outlier and hence making it very difficult for $D$ to overfit. To give a reference, we trained the exact same model without using the $G$ networks. This can be seen in Figure~\ref{simple_bon_pure}. We see that in the case with a feedforward neural network alone, the decision boundary for the red class is attempting to stretch towards the outlier, which is exactly avoided in the BON case. We will later do an analysis of the BON approach on a real life example with a more complex dataset, where we will also benchmark against using dropout.

Before we do so, it is important to note the negative evolution of the synthetic data points going from epoch $150$ to $500$ of the BON. Many of the synthetic data points which are correctly classified at epoch $150$ are suddenly collapsing to the real data point they were generated from. The reason for this is that the generator network only trains on misclassified samples at this stage. At epoch $150$ almost all samples are correctly classified, hence all that $G_k$ sees is the outlier. It keeps pushing the synthetic data point corresponding to this outlier close to the outlier itself (as we want it to), however in the process the neural network updates all of its weights such as to fulfill the objective of this single data point, hence forgetting how to generate the remaining data points. This is catastrophic forgetting, which is something neural networks are notoriously known for when switching from one task to another. The difference here is that we do not necessarily have a well-defined switch between tasks as is normally the case when academia try to tackle the issue. However, by making use of a per-epoch version of Memory Aware Synapses \cite{2017arXiv171109601A}, we manage to successfully overcome the issue as we will show on a more complex dataset.

\section{BON++}

In order to make the BON algorithm work, we need to make sure that the generator networks do not forget how to create synthetic data points for those data points that are correctly classified by $D$. The problem is that $G$ updates \textit{all} the weights of the network to alter the misclassified samples, hence removing the memory it has achieved for the correct samples. The way we will tackle this is by making approximations to the importance of the weights in $G$. If a particular weight is being used at a given time, we will make it harder for the network to update that particular weight. Instead we force $G$ to use other, less used, weights to learn new things.

\subsection{Memory Aware Synapses}

Memory Aware Synapses was recently introduced as an approach to \textit{learning what (not) to forget} \cite{2017arXiv171109601A}. In the following we will briefly describe the procedure as it is presented in the original article. The main idea is that we want to estimate parameter importance. Using the parameter importance, we can restrict the network to change the least important parameters to learn a new task and preserve the current most important parameters. In this way we prevent catastrophic forgetting of already learned tasks. 
Assume a single data point, $x^{(i)}$, the output of $G$ is $G(x^{(i)}; \theta_G)$ and if we change the parameters with a small amount $\delta$ the output is $G(x^{(i)}; \theta_G+\delta)$. Now, we can estimate the difference of the network output before and after by the gradients of the output w.r.t. each of the parameters, hence we have:
\begin{equation}
 G(x^{(i)}; \theta_G+\delta) - G(x^{(i)}; \theta_G)  \approx \sum_{i,\ j} g_{ij}(x^{(i)})\cdot \delta_{ij}
\end{equation}where $$g_{ij} = \frac{\partial G(x^{(i)}; \theta_G)}{\partial \left(\theta_{G}\right)_{ij}}$$
We can then use the value of $g_{ij}$ as an estimate of the parameter importance for the network $G$ on that particular data point $x^{(i)}$. In fact we can continuously update our measure of importance whenever the network sees a new data point, hence giving the following estimate of parameter importance for $m$ data points:
\begin{equation}
  \Omega_{ij} = \frac{1}{m} \sum_{i=1}^{m} \left\Vert\ g_{ij}\left(x^{(i)}\right)\right\Vert\
\end{equation}This in fact allows us to update the parameter importance whenever the network sees a new data point, i.e. in a streaming fashion. 
 
Since $G$ has more than one output, we estimate $g_{i j}$ as the $l^2$-norm of the outputs, i.e.
$$g_{i j} = \frac{\partial\ l^2 \left[ G(x^{(i)}; \theta_G)\right]}{\partial \left(\theta_{G}\right)_{ij}}$$Having calculated the parameter importance as above, we need to incorporate this in the loss function of $G$. In the original article, they fix a set of parameters $\theta^{*}$after each task. In our setting, $G$ does not have specific tasks, instead we fix the network parameters after each epoch and denote these $\theta^{*}$. Given that we are only showcasing small datasets in this paper, it may be more beneficial to fix the parameters more often for larger datasets. Having fixed the parameters, the loss is then calculated as the distance from the current parameters to these fixed parameters multiplied by how important the parameters are and the total loss for $G$ is hence:

\begin{equation}
\resizebox{.9\hsize}{!}{$L(\theta_G) = \alpha \cdot \underbrace{MSE\left(\hat x^{(i)}_M, x^{(i)}\right)}_{\text{original loss}} + \beta^{n} \cdot \underbrace{\sum_{i,j} \Omega_{ij} \left( \left(\theta_{G}\right)_{ij} - \left(\theta_{G}^*\right)_{ij}\right)^2}_{\text{Memory Aware Synapses loss}}$} 
\end{equation}
where $n$ is the number of epochs, $\alpha$ and $\beta > 1$ are hyperparameters. The reason we raise the hyperparameter $\beta$ to the number of epochs is that we don't want to place too much emphasis on the parameter importance early on in training. As training progresses, we want to memorize more and more of the important parameters since more and more data points will be classified correctly.

\subsection{BON++ Algorithm}
The final BON algorithm, including Memory Aware Synapses to prevent catastrophic learning, is given in Algorithm~\ref{alg:bon++}.

 \begin{algorithm}[h!]
   \caption{BON++}
   \label{alg:bon++}
\begin{algorithmic}
   \STATE {\bfseries Input:} dataset $(x^{(i)}, y^{(i)}), i \in \{1,2,...,m\}$
   \STATE
   \FOR{$n=1$ {\bfseries to} $\# \ epochs$}
	\FOR{$k=1$ {\bfseries to} $K$}   
    \STATE
   \STATE $ \left(\theta_{G_k}^*\right)_{ij} := \left(\theta_{G_k}\right)_{ij}, \quad \forall i,j $
	\STATE
    \ENDFOR
	\STATE
	\FOR{$i=1$ {\bfseries to} $m$}
	\FOR{$k=1$ {\bfseries to} $K$}   
    \STATE
   \STATE $\bullet$ Create synthetic data point $G_k(x^{(i)}) = \hat x^{(i)}_k$   
   \STATE
   \STATE $\bullet$ Compute prediction of synthetic data point: \\
   \STATE
   \STATE   $\quad \quad \quad \quad \quad \quad D(\hat x^{(i)}_k) = \hat y^{(i)}_{\hat x_k}$      
      \ENDFOR
   \STATE
	\STATE $\bullet$ Compute prediction of real data point:
    \STATE
    $\quad \quad \quad \quad \quad \quad D(x^{(i)}) = \hat y^{(i)}_{x}$ 
    \STATE
	\STATE $\bullet$ Compute loss for $D$: \\
   \STATE
	\STATE    \resizebox{.9\hsize}{!}{$L(\theta_D)  = \sum_{k} CE\left[D\left(G_k\left(x^{(i)}\right)\right),  y^{(i)}\right]+ CE\left[\hat y^{(i)}_{x},  y^{(i)}\right]$}   
   \STATE    
	\STATE $\bullet$ Update $\theta_D$ to minimize $L(\theta_D)$ 
   \STATE
	\FOR{$k=1$ {\bfseries to} $K$}   
   \STATE $\bullet$ Calculate importance weights:  
    \STATE
	\STATE \resizebox{.7\hsize}{!}{$\quad \quad \quad \quad \quad \quad g_{i j} = \frac{\partial\ l^2 \left[ G_k(x^{(i)}; \theta_{G_k})\right]}{\partial \left(\theta_{G_k}\right)_{ij}}, \quad \forall i,j$}  
   \ENDFOR
   \STATE
\IF{$\hat c \neq y^{(i)}$}
   \FOR{$k=1$ {\bfseries to} $K$} 
    \STATE
	\STATE $\bullet$ Compute MSE loss for $G_k$:
    \STATE
    \STATE \quad $L_1(\theta_{G_k}) = \alpha \cdot MSE\left[G_k\left(x^{(i)}\right), x^{(i)}\right]$ 
   	\STATE
   \STATE $\bullet$ Compute MAS loss for $G_k$:
    \STATE
    \STATE \resizebox{.7\hsize}{!}{$L_2(\theta_{G_k}) = \beta^{n} \cdot \sum_{i,j} \Omega_{ij} \left( \left(\theta_{G_k}\right)_{ij} - \left(\theta_{G_k}^*\right)_{ij}\right)^2$}    
	 \STATE
    \STATE $\bullet$ Update $\theta_{G_k}$ to minimize $L_1(\theta_{G_k})+L_2(\theta_{G_k})$    
   \STATE
         \ENDFOR
      \ENDIF
   \ENDFOR
   \ENDFOR
\end{algorithmic}
\end{algorithm}

\section{Experiments: BON++}

Having established the algorithm, we will now demonstrate it on a real life scenario. Concretely, we choose the Iris dataset where we use the sepal length and width as input features. In this 2D space, one class is easily separable, but the two other classes are tough to separate. We first show the baseline using no dropout and then using dropout of 20\%, 50\% and 80\%. We do so to establish a benchmark, since there is no right or wrong when deciding which decision boundary is better than the other. However, we will be able to visually show differences and then conclude whether BON++ is doing what we expect.
\subsection{Real example: Iris dataset }

In the Iris experiment \cite{AHG:AHG2137}, we use the same architectures for both $D$ and $G$ as in the constructed example, except that $D$ has $3$ output neurons to match that we have $3$ classes in the Iris dataset\footnote{The experiment is done in PyTorch and will be released soon.}. 

We use the following hyperparameters: \\
$\bullet$ $K = 100$\\
$\bullet$ batch size $=10$ \\
$\bullet$ $\alpha = 2$\\
$ \bullet$ $\beta = 1.025$ \\
$\bullet$ $lr_D = 0.001$\\
$\bullet$ $lr_G = 0.0001$\\

\begin{figure}[H]

\centering
\includegraphics[trim= 0cm 0cm 0cm 0.7cm, scale=0.40]{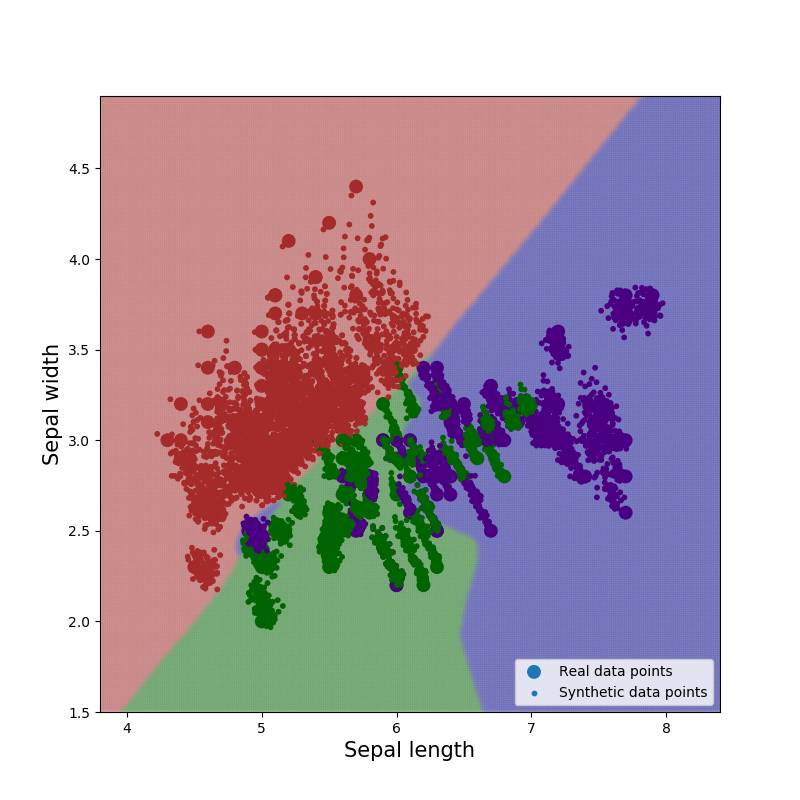}

\caption{Iris dataset: BON++ algorithm after 1,000 epochs. The colors represent the hard decision boundaries, i.e. the class prediction, not to be confused with the actual probabilities.}
\label{bon_iris}

\end{figure}

\begin{figure}[h!]
\centering
\subfigure[Dropout $= 0.0$]{\label{fig:a}\includegraphics[width=45mm]{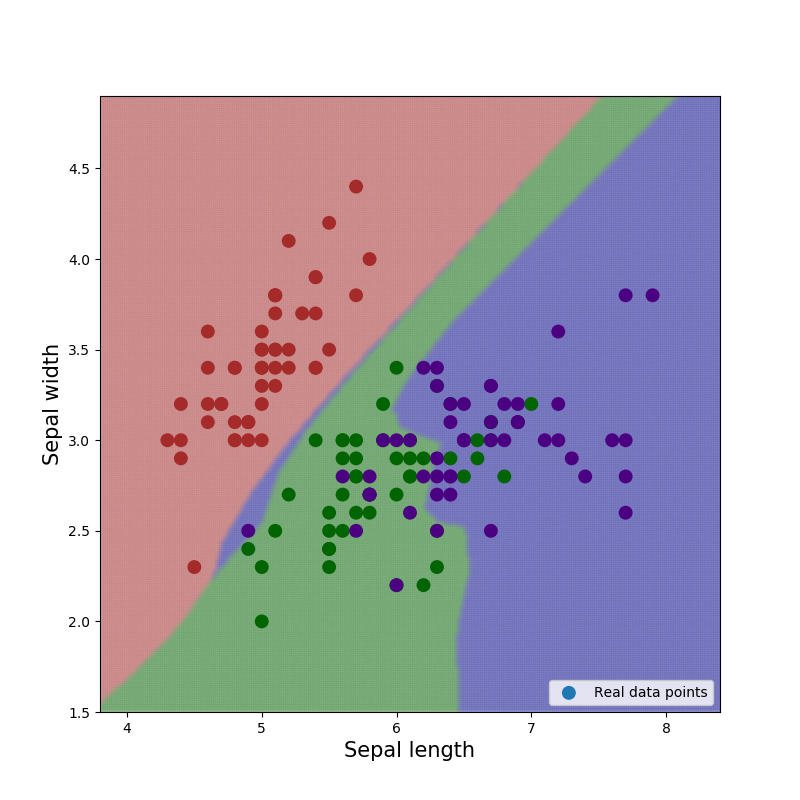}}
\subfigure[Dropout $= 0.2$]{\label{fig:b}\includegraphics[width=45mm]{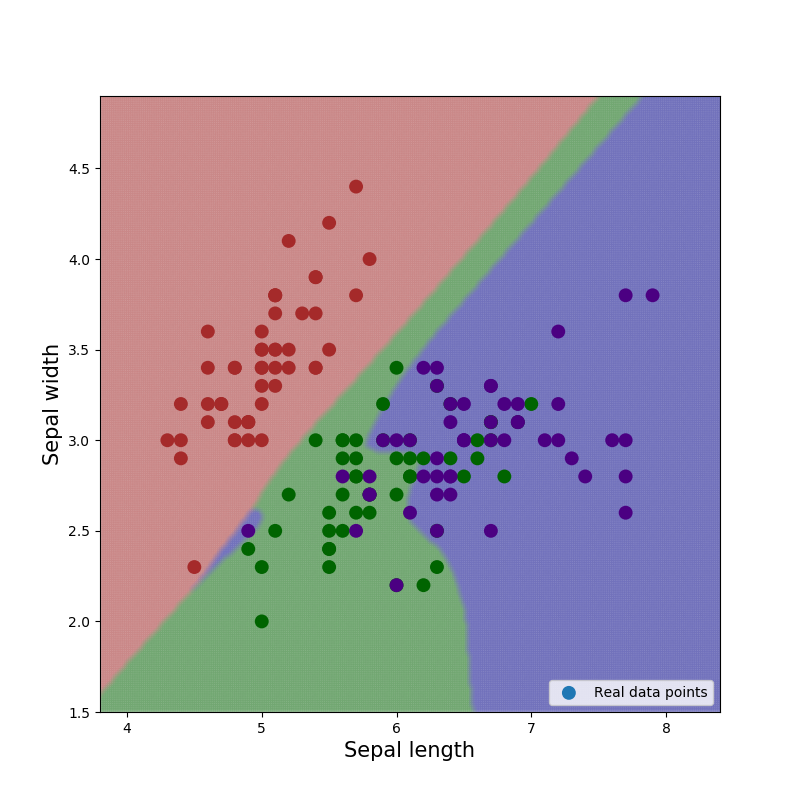}}
\subfigure[Dropout $= 0.5$]{\label{fig:c}\includegraphics[width=45mm]{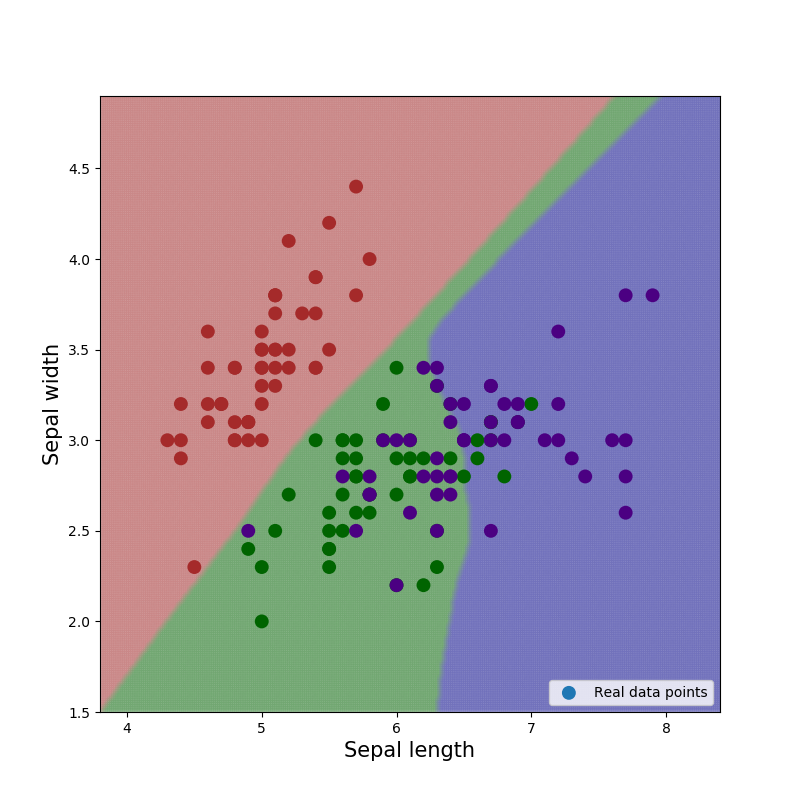}}
\subfigure[Dropout $= 0.8$]{\label{fig:d}\includegraphics[width=45mm]{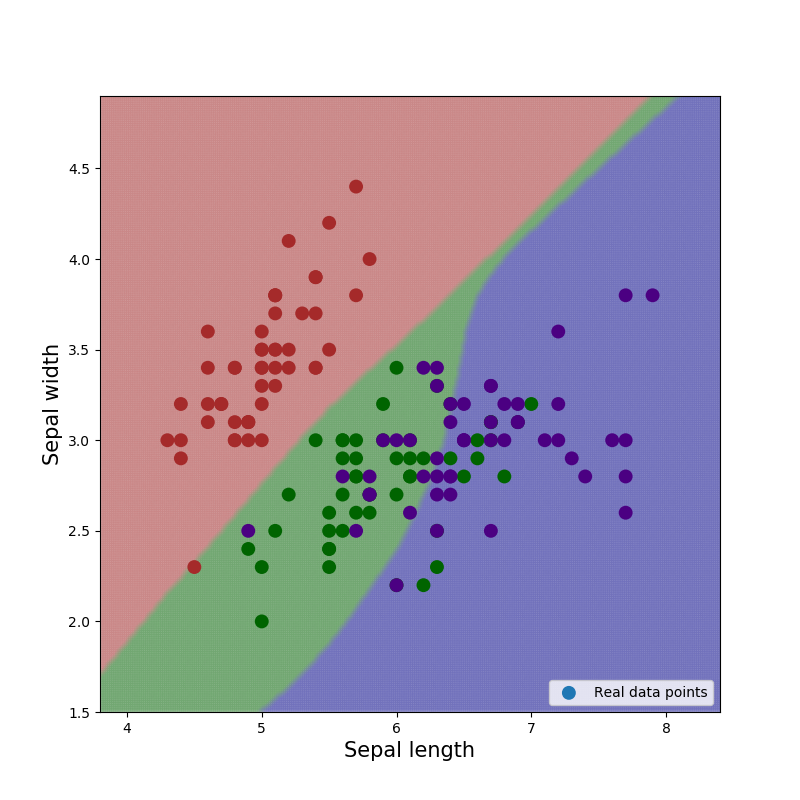}}
\caption{Iris dataset: Feedforward neural network at different levels of dropout after $1,000$ epochs. The colors represent the hard decision boundaries, i.e. the class prediction, not to be confused with the actual probabilities.}
\label{simple_iris}
\end{figure}

Figure \ref{bon_iris} shows the result of the BON++ algorithm and figure \ref{simple_iris} shows the feedforward neural network for different levels of dropout.

As expected, BON++ is creating support for existing data points, hence blocking for potential overfitting in most areas. We particularly see that for the large green area in the top-right of figure \ref{simple_iris} (a)-(d), which is prevented by the large mass of supporting synthetic data points in figure  \ref{bon_iris}. 

Furthermore we see that BON++ preserves some of the expressiveness w.r.t. the decision boundary between the green and purple class, i.e. when the data points are mixed in a complex manner the approach is still capable of creating a complex decision boundary. On the other hand, using dropout creates a clear trade-off between simplicity of decision boundary and overfitting, hence we cannot prevent overfitting in some areas while preserving complexity in other areas. 

The per-epoch version of Memory Aware Synapses is working as intended, although the model trains for many epochs ($1,000$), the supporting data points in areas where $D$ correctly classifies the synthetic data points early on in training are \textit{not} suffering from catastrophic forgetting anymore. Memory Aware Synapses played a key role in the robustness of the algorithm.

\section{Discussion \& Conclusion}
In this paper we introduced a novel approach to generalization of neural networks. 
We showed that the method had an initial positive effect during training of CIFAR-10 using Densenet. However as training progressed, the generator network suffered from catastrophic forgetting leading to the collapse of all synthetic data points to the real equivalent it was generated from.
We furthermore showed that the method works on a simple constructed data set for which the BON algorithm were able to prevent overfitting to a single outlier. 

We then developed the BON++ algorithm using a per-epoch version of Memory Aware Synapses and tested it on the Iris dataset, where the decision boundary is much harder for two of the classes than in the constructed example. The model did prevent most overfitting while keeping some expressiveness in the difficult areas. 

Due to the late discovery of Memory Aware Synapses we have not completed a full study of BON++ on larger datasets, hence for future work we will adopt the approach to CIFAR-10 and other domains with a higher-dimensional input space.
We will also make the code used in the experiments public such that people in different domains can test it.

The BON++ approach do increase the training time significantly, however it is worth mentioning that the number of generators can be run in parallel since they are each, independently, creating synthetic data points. The added training time of this training approach is hence only in the order of training time for a single $G$. Although the training time is increased, there is no impact on speed during inference, since the classifier is equal to the classifier in any other paradigm. 

This leads to the next point; the BON++ approach works for any architecture of $D$. The only requirement is to be able to create a generator that effectively can create synthetic data points, while taking as inputs the real data points.

One of the areas we did not test yet is to use a different measure of similarity for two data points. We currently use the mean squared error between the synthetic data point and its real equivalent, however one could use different measures. One such measure is to push the synthetic data point closer to an intermediate layer of $D$ evaluated on the real data point. This could potentially push data points closer to an internal representation of the real data points, hence creating more real life augmentations than the current approach. 

\clearpage
\section*{Acknowledgements}
We thank Lars Maal{\o}e (Corti), Tycho Tax (Corti), Thomas Jakobsen (Grazper) and Stefan Sommer (University of Copenhagen, Department of Computer Science) for valuable comments and ideas.

\bibliography{example_paper}
\bibliographystyle{icml2018}


\end{document}